\documentclass[final]{cvpr}

\usepackage{times}
\usepackage{epsfig}
\usepackage{graphicx}
\usepackage{amsmath}
\usepackage{amssymb}
\usepackage{subfigure}
\usepackage{booktabs}
\usepackage{multirow}
\usepackage[misc]{ifsym}
  
\usepackage[pagebackref=true,breaklinks=true,colorlinks,bookmarks=false]{hyperref}

\def\cvprPaperID{10472} 
\def\confYear{CVPR 2021}

\begin{document}

\title{MoNet: Motion-based Point Cloud Prediction Network}

\author{Fan Lu\textsuperscript{1},\quad Guang Chen\textsuperscript{1,2,\Letter},\quad Yinlong Liu\textsuperscript{2},\quad Zhijun Li\textsuperscript{3}, \quad Sanqing Qu\textsuperscript{1}, \quad Tianpei Zou\textsuperscript{1}\\
\textsuperscript{1}Tongji University, \textsuperscript{2}Technische Universität München, \\ \textsuperscript{3}University of Science and Technology of China\\
{\tt\small lufan@tongji.edu.cn, \quad guangchen@tongji.edu.cn, \quad Yinlong.Liu@tum.de}\\
{\tt\small zjli@ieee.org, \quad 2011444@tongji.edu.cn, \quad 2011459@tongji.edu.cn}
}

\maketitle


\begin{abstract}
     Predicting the future can significantly improve the safety of intelligent vehicles, which is a key component in autonomous driving. 3D point clouds accurately model 3D information of surrounding environment and are crucial for intelligent vehicles to perceive the scene. Therefore, prediction of 3D point clouds has great significance for intelligent vehicles, which can be utilized for numerous further applications. However, due to point clouds are unordered and unstructured, point cloud prediction is challenging and has not been deeply explored in current literature. In this paper, we propose a novel motion-based neural network named MoNet. The key idea of the proposed MoNet is to integrate motion features between two consecutive point clouds into the prediction pipeline. The introduction of motion features enables the model to more accurately capture the variations of motion information across frames and thus make better predictions for future motion. In addition, content features are introduced to model the spatial content of individual point clouds. A recurrent neural network named MotionRNN is proposed to capture the temporal correlations of both features. Besides, we propose an attention-based motion align module to address the problem of missing motion features in the inference pipeline. Extensive experiments on two large scale outdoor LiDAR datasets demonstrate the performance of the proposed MoNet. Moreover, we perform experiments on applications using the predicted point clouds and the results indicate the great application potential of the proposed method.
\end{abstract}

\section{Introduction}

Predicting the future is an essential capability of intelligent vehicles, which allows for anticipation of what will happen in the future and can significantly improve the safety of autonomous driving \cite{rudenko2020human,mozaffari2020deep,zhang2020stinet,kwon2019predicting,chen2020long}. 3D point clouds can provide accurate 3D modeling of the surrounding environment and have been widely used in numerous perception applications (\emph{e.g.}, object detection \cite{shi2020pv,shi2019pointrcnn,yan2018second,yang20203dssd}, semantic segmentation \cite{hu2020randla,milioto2019rangenet++,wu2018squeezeseg,zhang2020polarnet}). Compared to existing video prediction \cite{jin2020exploring,kwon2019predicting,wang2018eidetic,wang2017predrnn}, the prediction of 3D point clouds can provide an opportunity for a better understanding of future scenes \cite{guo2020deep}, which can help the intelligent vehicles perceive the environment and make decisions in advance. However, 3D point clouds are unordered and unstructured \cite{qi2017pointnet}, which makes the prediction of 3D point clouds more challenging than video prediction and there is little exploration of point cloud prediction in the current literature.

Based on the above observations, we study a problem named \textit{Point Cloud Prediction} in this paper. Concretely, given a sequence contains $T$ frames of point clouds $\{P_1,\cdots,P_{T}\}$, point cloud prediction aims to predict $T_p$ future frames $\{P_{T+1},\cdots,P_{T+T_p}\}$. The prediction of future point clouds can be formulated as a per-point motion prediction problem. Intuitively, the prediction model should be capable of capturing the variations of motion information across point cloud frames, which can benefit from explicitly estimating motion features between two point clouds. However, the introduction of motion features in point cloud prediction has not been explored in existing methods \cite{fan2019pointrnn, Weng2020_SPF2}.

To address the above problem, we propose a motion-based neural network named MoNet for point cloud prediction. Different from previous works \cite{fan2019pointrnn, Weng2020_SPF2}, the key idea of MoNet is to integrate motion information between two consecutive point clouds into point cloud prediction. Motion features between two point clouds are extracted using a motion encoder, which allows the network to accurately capture the variations of motion information. Content features of an individual frame are introduced to model the spatial content of a point cloud itself, which helps preserve spatial structures of point clouds. The combination of content and motion features can exploit complementary advantages and results in better performance. To capture the temporal correlations of both features across frames, we propose a recurrent neural network (RNN) named MotionRNN. Unlike standard RNN and its variants like Long Short-Term Memory (LSTM) \cite{hochreiter1997long} and Gated Recurrent Unit (GRU) \cite{cho2014learning} which can only process one-dimensional features, the proposed MotionRNN associates the states with the coordinates of points to maintain the spatial structure of point clouds. During the inference pipeline, an attention-based motion align module is proposed to estimate the motion features to address the absence of the next point cloud. The overall architecture of the proposed MoNet is shown in Fig.~\ref{fig:network}.

To verify the feasibility of the proposed method, we perform extensive experiments on two large scale outdoor LiDAR point cloud datasets, namely KITTI odometry dataset \cite{geiger2012we} and Argoverse dataset \cite{chang2019argoverse}. Both qualitative and quantitative results show that the proposed MoNet significantly outperforms baseline methods and is capable of effectively predicting the future point clouds. Besides, the experiments on applications indicate the great application potential of the proposed method.

To summarize, our main contributions are as follows:
\begin{itemize}
\setlength{\itemsep}{0pt}
\setlength{\parsep}{0pt}
    \item A novel motion-based neural network named MoNet is proposed for point cloud prediction, which can effectively predict the future point clouds.
    \item Motion features are explicitly extracted and integrated into point cloud prediction and the combination of motion and content features results in better performance. Besides, MotionRNN is proposed to model the temporal correlations of both features.
    \item An attention-based motion align module is proposed in the inference pipeline to estimate the motion features without future point clouds.
\end{itemize}

\section{Related works}
We briefly review the most related works, including video prediction, sequential point clouds processing and point cloud prediction.

\textbf{Video prediction} Video prediction aims to predict future images given a sequence of previous frames. ConvLSTM \cite{xingjian2015convolutional} introduced 2D convolutions into LSTM to extract visual representations, which has been a seminal work in video prediction. Based on \cite{xingjian2015convolutional}, Wang et al. \cite{wang2017predrnn} proposed ST-LSTM to memorize spatial appearance and temporal variations simultaneously. Video pixel network (VPN) \cite{kalchbrenner2017video} utilized PixelCNNs to directly estimate the discrete joint distribution of the raw pixel values for video prediction. Villegas et al. \cite{villegas17mcnet} proposed to decompose motion and content and independently capture each information stream.  Eidetic 3D LSTM (E3D-LSTM) \cite{wang2018eidetic} was proposed to integrate 3D convolutions into RNNs for video prediction and exploit complementary advantages. Generative adversarial network (GAN) has also been widely applied in video prediction to improve the quality of generated images \cite{liang2017dual, kwon2019predicting}.

\textbf{Sequential point clouds processing} Processing sequential point clouds is challenging due to that point clouds are disordered and unstructured. FlowNet3D \cite{liu2019flownet3d} utilized flow embedding layer to model the correspondence of points between two point clouds, which is a representative work to process consecutive point clouds using deep learning-based method. Liu et al. \cite{liu2019meteornet} proposed MeteroNet, which learns to aggregate information from spatiotemporal neighbor points to learn representations for dynamic point cloud sequences. In \cite{shi2020spsequencenet}, a cross-frame global attention module and local interpolation module were proposed to capture spatial and temporal information in point cloud sequences for point cloud semantic segmentation. In \cite{min2020efficient}, a weight-shared LSTM model named PointLSTM was proposed to update state information for neighbor point pairs to perform gesture recognition. Choy et al. \cite{choy20194d} proposed 4D spatio-temporal convolutional network for 3D-video perception. 

\textbf{Point cloud prediction} Point cloud prediction has not been deeply explored in literature. Fan et al. \cite{fan2019pointrnn} proposed PointRNN to model temporal information across point clouds for future point clouds prediction. They proposed a point-based spatiotemporal-local correlation named point-rnn to replace concatenation operation in standard RNN to process moving point clouds. Very recently, Weng et al. \cite{Weng2020_SPF2} adopted an encoder-decoder architecture to predict future point clouds and utilizes the predicted point clouds to perform trajectory prediction.

\section{Methodology}
\begin{figure*}
    \centering
    \includegraphics[width=0.99\textwidth]{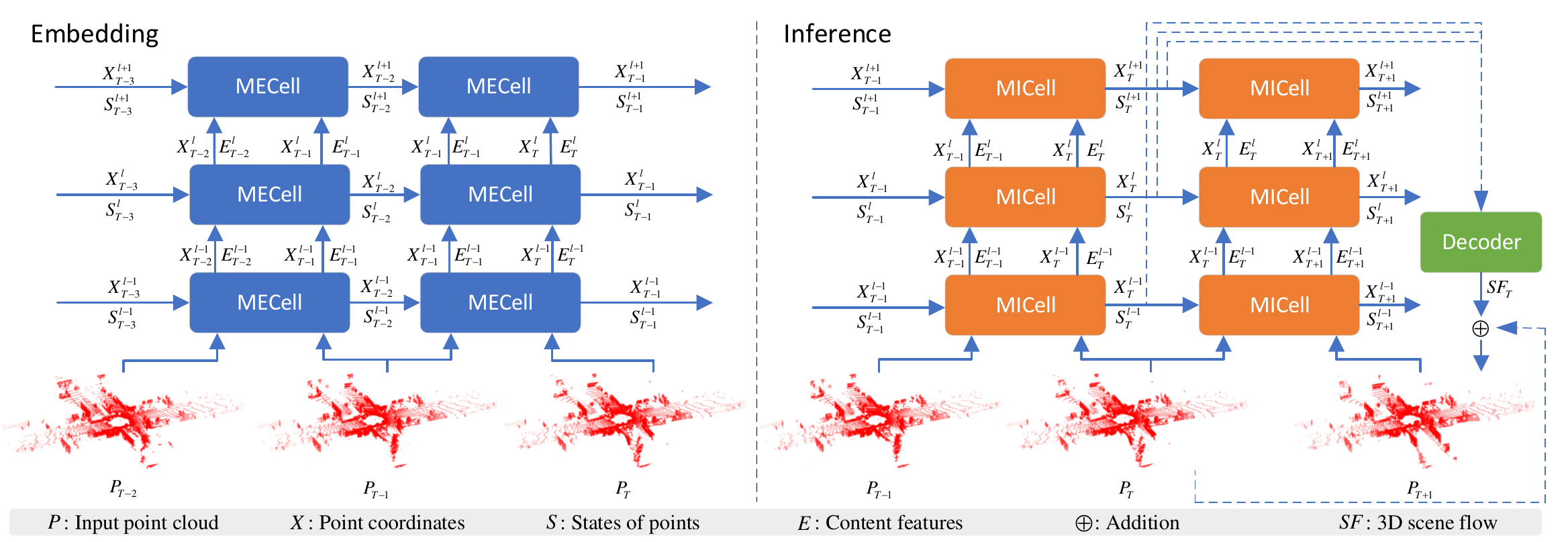}
    \caption{Overall architecture of the proposed MoNet. The left part displays the embedding pipeline, where the MECell (Motion Embedding Cell) firstly extracts content features and motion features and then captures temporal information across point clouds. The right part shows the inference pipeline and the MICell (Motion Inference Cell) is adopted to replace MECell to predict future frames.}
    \label{fig:network}
\end{figure*}

\subsection{Overall architecture}
Given a point cloud sequence $\{P_1,\cdots,P_{T}\}\in\mathbb{R}^{N\times 3}$, the goal of point cloud prediction is to predict $T_p$ future point clouds $\{P_{T+1},\cdots,P_{T+T_p}\}\in\mathbb{R}^{N\times 3}$. To achieve that, we propose a novel motion-based neural network named MoNet and the overall architecture is shown in Fig.~\ref{fig:network}. MoNet can be divided into two pipelines, namely embedding pipeline and inference pipeline. The input point clouds are firstly inputted into the embedding pipeline to extract the content and motion features and capture the temporal correlations across point clouds. Then the embedded features are passed into the inference pipeline to predict the future point cloud frames. We denote the point coordinates, content features and motion features of frame $t$ in layer $l$ as $X_t^l\in \mathbb{R}^{N_l\times 3}$, $E_t^l\in \mathbb{R}^{N_l\times d_l^E}$ and $M_t^l\in \mathbb{R}^{N_l\times d_l^M}$, where $N_l$ is the number of points in layer $l$, $d_l^E$ and $d_l^M$ are number of channels of content and motion features, respectively. We will describe the two pipelines in detail below.

\subsection{Embedding}
\begin{figure}
    \centering
    \includegraphics[width=0.45\textwidth]{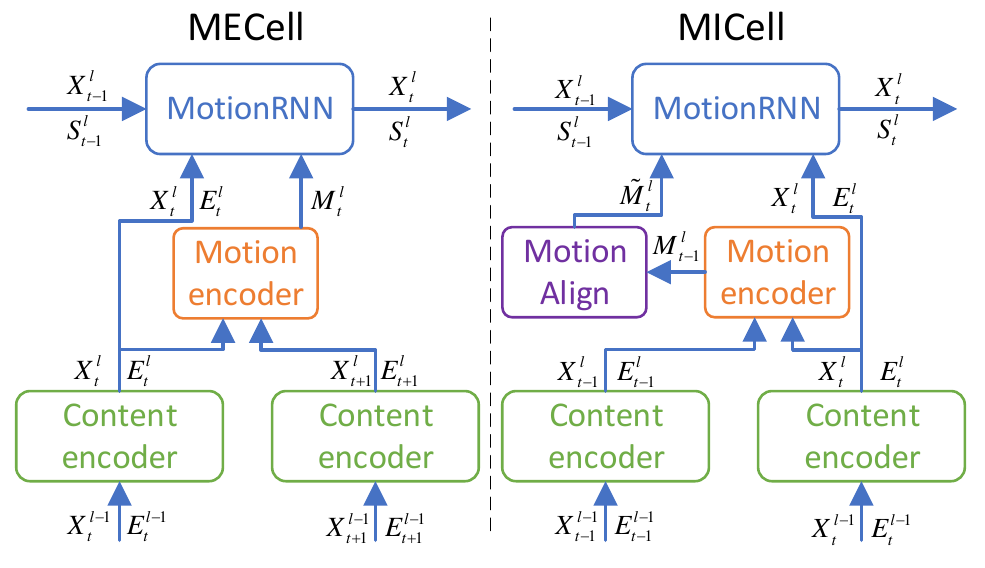}
    \caption{Left: MECell (Motion Embedding Cell). Right: MICell (Motion Inference Cell).}
    \label{fig:motioncell}
    \vspace{-5mm}
\end{figure}

As shown in the left part of Fig.~\ref{fig:network}, the key component of the embedding pipeline is the MECell (\emph{i.e.}, Motion Embedding Cell). As we mentioned before, content features of an individual frame and motion features of consecutive point clouds can both contribute to point cloud prediction. However, previous works only model the temporal information of point cloud sequence and do not explicitly extract the motion features between point clouds. As shown in the left part of Fig.~\ref{fig:motioncell}, given content features and point coordinates $(X_t^{l-1},E_t^{l-1})$, $(X_{t+1}^{l-1},E_{t+1}^{l-1})$ of two consecutive point clouds from the last layer $l-1$, the content encoder is firstly adopted to extract point coordinates and content features $(X_t^{l},E_t^{l})$, $(X_{t+1}^{l},E_{t+1}^{l})$ of current layer $l$. Then motion encoder is followed to generate motion features $M_t^{l}$ corresponding to frame $t$, which models the per-point motion information from frame $t$ to frame $t+1$. After that, $M_t^l$, $E_t^l$, $X_t^l$, the states $S_{t-1}^l$ and coordinates $X_{t-1}^l$ from last frame $t-1$ are passed into a recurrent neural network named MotionRNN to simultaneously capture the temporal information of motion features and content features across point clouds and generate the states $S_t^l$ of current frame $t$. 

\subsubsection{Content encoder}
We adopt a PointNet++\cite{qi2017pointnet++}-like structure to encode the content features of point clouds. Given $X_t^{l-1}$ and $E_t^{l-1}$ from the last layer, firstly $N_{l}$ points $X_t^l$ are sampled from $X_t^{l-1}$ using Furthest Point Sampling (FPS). For each point $x_i \in X_t^{l}$, $k$ neighbor points $\{x_i^1,\cdots,x_i^k\}$ are searched in $X_t^{l-1}$ around $x_i$ using $k$-nearest-neighbors ($k$NN) method to generate a cluster. The relative coordinates $\{x_i^1-x_i,\cdots,x_i^k-x_i\}$ and the relative distances $\{\left\|x_i^1-x_i\right\|, \cdots,\left\|x_i^k-x_i\right\|\}$ are calculated as geometric features of the cluster, where $\left\|\cdot\right\|$ denotes Euclidean distance. The geometric features are then concatenated with content features $\{e_i^1,\cdots,e_i^k\}$ to generate a feature map, where $e_i^j$ is the content feature corresponding to $x_i^j$. After that, the feature maps of all clusters are passed into Shared Multilayer Perceptron (Shared-MLP) with maxpool layer to generate content features $E_{t}^l$ of the current layer $l$.

\subsubsection{Motion encoder}
Motion encoder is proposed to model the motion information between two consecutive point clouds. The input of motion encoder are the point coordinates and content features $(X_t^l,E_t^l)$, $(X_{t+1}^l,E_{t+1}^l)$ of two point clouds, and the output is the motion features from $X_t^l$ to $X_{t+1}^l$. For each point $x_i$ in $X_{t}^l$, we search $k$ neighbor points in $X_{t+1}^l$ and then use a similar strategy as content encoder to generate geometric features. The feature map is a concatenation of geometric features, content features of neighbor points in $X_{t+1}^l$ and content features of the center point $x_i$. After that, Shared-MLP with maxpool layer are followed to generate motion features $M_t^l$. Intuitively, the content features of two point clouds and geometric features between two frames all contribute to the generated motion features, which enables the motion encoder to capture the changes of contents and point coordinates across two point clouds. Thus, motion features of frame $t$ can be considered as a representation of motion information from frame $t$ to $t+1$.

\subsubsection{MotionRNN}
General RNN models like LSTM and GRU can only process one-dimensional features. However, spatial information is important in the representation of point clouds. A one-dimensional global feature is hard to preserve the spatial structure and local details of point clouds thus is not applicable for large scale point cloud prediction. Based on the above consideration, we propose a novel recurrent neural network named MotionRNN for point cloud prediction. The key idea of MotionRNN is that the states correspond to point coordinates one by one and the content and motion features are combined to update the states. The inputs of MotionRNN are the points $X_{t-1}^l$ and states $S_{t-1}^l$ from last frame, the points $X_t^l$, content features $E_{t}^{l}$ and motion features $M_t^l$ of current frame and the output are the points $X_{t}^l$ and states $S_{t}^l$ of current frame. 

We provide two versions of MotionRNN, namely MotionGRU and MotionLSTM. Taken MotionLSTM as an example, the states $S_t^l$ consists of hidden states $H_t^l\in \mathbb{R}^{N_l\times d_l^S}$ and cell states $C_t^l\in \mathbb{R}^{N_l\times d_l^S}$. For each point in $X_t^l$, $k$ neighbor points are searched in $X_{t-1}^l$ and the similar strategy in content encoder is adopted to generate the geometric features $\Bar{G}_t^l$ and also clusters. Denote the hidden state and cell state of clusters as $\Bar{H}_{t-1}^l$ and $\Bar{C}_{t-1}^l$, then the updates for MotionLSTM at $t$-th step in layer $l$ can be formulated as:

\begin{equation}
    \begin{split}
       &I_t^l = \sigma\left({\rm maxpool}({\rm MLP}([\Bar{G}_t^l,\Bar{H}_{t-1}^l,M_t^l,E_t^l]))\right) \\
       &F_t^l = \sigma\left({\rm maxpool}({\rm MLP}([\Bar{G}_t^l,\Bar{H}_{t-1}^l,M_t^l,E_t^l]))\right) \\
        &O_t^l = \sigma\left({\rm maxpool}({\rm MLP}([\Bar{G}_t^l,\Bar{H}_{t-1}^l,M_t^l,E_t^l]))\right) \\
        &\hat{C}_{t-1}^l = {\rm maxpool} ({\rm MLP}([\Bar{G}_t^{l},\Bar{C}_{t-1}^l])) \\
        &\tilde{C}_t^l = {\rm tanh}\left({\rm maxpool}({\rm MLP}([\Bar{G}_t^l,\Bar{H}_{t-1}^l,M_t^l,E_t^l]))\right) \\
        &C_t^l = F_t^l \odot \hat{C}_{t-1}^l + I_t^l \odot \tilde{C}_t^l \\
        &H_t^l = O_t^l \odot {\rm tanh} (C_t^l)
    \end{split}
\end{equation}
where $\sigma(\cdot)$, $[\cdot]$ and $\odot$ represent Sigmoid function, concatenation operation and Hadamard product, respectively. Similarly, the updates for MotionGRU can be represented as Eq.~\ref{eq:motiongru}, where the states $S_t^l$ consists of only hidden states $H_t^l$.

\begin{equation}
\label{eq:motiongru}
    \begin{split}
        &Z_t^l=\sigma\left({\rm maxpool}({\rm MLP}([\Bar{G}_t^l,\Bar{H}_{t-1}^l,M_t^l,E_t^l]))\right)\\
        &R_t^l=\sigma\left({\rm maxpool}({\rm MLP}([\Bar{G}_t^l,\Bar{H}_{t-1}^l,M_t^l,E_t^l]))\right)\\
        &\hat{H}_{t-1}^l={\rm maxpool} ({\rm MLP}([\Bar{G}_t^{l},\Bar{H}_{t-1}^l]))\\
        &\tilde{H}_t^l={\rm tanh}({\rm MLP}([R_t^l\odot \hat{H}_{t-1}^l,M_t^l,E_t^l]))\\
        &H_t^l=Z_t^l\odot \hat{H}_{t-1}^l+(1-Z_t^l)\odot \tilde{H}_t^l
    \end{split}
\end{equation}

Compared to standard RNN models, the proposed MotionRNN corresponds states to coordinates one by one. This representation can well maintain the spatial structure of point clouds, thus is more applicable for point cloud prediction. Different from PointRNN \cite{fan2019pointrnn}, the motion features and content features are incorporated into the recurrent neural network, which enables the network to simultaneously capture the variations of motion information and temporal correlations of spatial contents across frames.

\subsection{Inference}
In the embedding pipeline, the motion features and content features are extracted and the temporal correlations are modeled using the proposed MotionRNN. In general prediction models, the process in the inference pipeline is exactly the same as that in the embedding pipeline, where the output states of the last frame of the embedding pipeline are duplicated to the first cell of the inference pipeline to predict the first future frame \cite{srivastava2015unsupervised,xingjian2015convolutional}. However, this general strategy is not applicable to our method. Noting that the motion features of frame $t$ are generated using the content features of frame $t$ and $t+1$. However, the next frame $T+1$ does not exist for the last frame $T$ of the embedding pipeline so that the motion features of frame $T$ is not available. 

To address the above problem, we propose MICell (\emph{i.e.}, Motion Inference Cell) in the inference pipeline to replace the MECell in the embedding pipeline. The architecture of MICell can be seen in the right part of Fig.~\ref{fig:motioncell}. Given $(X_{t-1}^{l-1}, E_{t-1}^{l-1})$ and $(X_{t}^{l-1}, E_{t}^{l-1})$, the content encoder and motion encoder are the same as that in MECell. However, to overcome the lack of motion features corresponding to frame $t$, a motion align module is applied after the motion encoder to align the motion features of frame $t-1$ to frame $t$, which is the only difference between MECell and MICell. Then the estimated motion features $\tilde{M}_{t}^l$ and content features $E_{t}^l$ with coordinates $X_{t}^l$ are input into MotionRNN to generate the states $S_{t}^l$. Finally, the decoder is followed to decode the hidden states of frame $t$ to the 3D scene flow $SF_{t}$ of points in $P_{t}$ and the future point cloud $P_{t+1}$ can be calculated as $P_{t+1}=P_{t}+SF_{t}$. Here we adopt the Feature Propagation module in PointNet++ \cite{qi2017pointnet++} as the decoder and decode the hidden states in a coarse-to-fine manner. 

\subsubsection{Motion align}
Motion align module is introduced to estimate motion features $\tilde{M}_{t}^l$ of current frame $t$ from the motion features of last frame $t-1$. Noting that the motion features are corresponding to the points coordinates one-by-one. To estimate the motion features, we make two basic assumptions based on the fact that the time between two consecutive point clouds is small: (1) From frame $t-1$ to frame $t$, the movement of a point is limited; (2) From frame $t-1$ to frame $t$, the motion feature of a point will not change significantly. Based on the above two assumptions, the motion feature of a point in the current frame can be estimated given the motion features of its neighbor points in the previous frame. 

\begin{figure}
    \centering
    \includegraphics[width=0.45\textwidth]{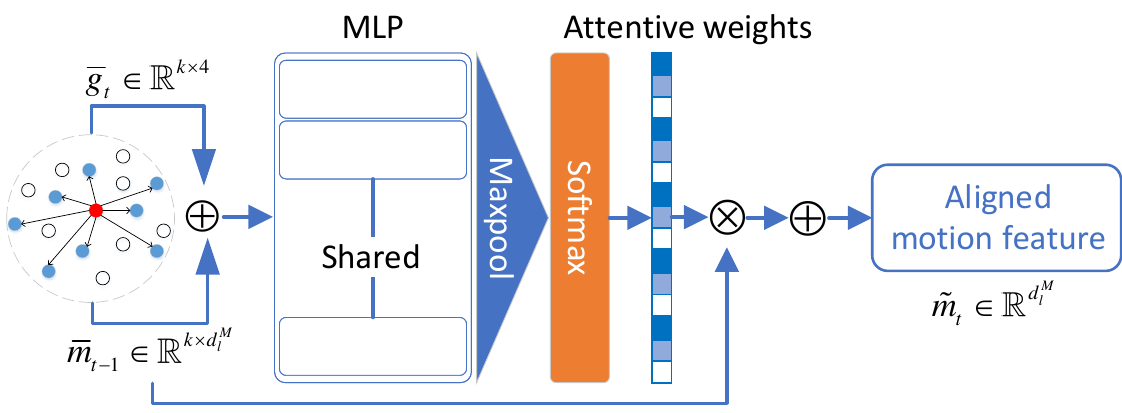}
    \caption{The architecture of the proposed motion align module. The left circle represents a $k$-nearest-neighbors cluster, where the red point is a point in frame $t$, blue points are neighbor points in frame $t-1$ and white points represent other points in frame $t$.}
    \label{fig:motionalign}
    \vspace{-4mm}
\end{figure}

Here we adopt an attention mechanism to perform the estimation of motion features. The network architecture of the proposed motion align module is shown in Fig.~\ref{fig:motionalign}. Taken a point $x_{t}^l$ in $X_{t}^l$ as an example, $k$ neighbor points are searched in $X_{t-1}^l$ and the geometric features $\bar{g}_{t}^l\in \mathbb{R}^{k\times 4}$ are extracted using the same operation as described in the content encoder. The geometric features are then concatenated with motion features $\bar{m}_{t-1}^l\in \mathbb{R}^{k\times d_l^M}$ of the neighbor points to produce a feature map. Shared-MLP and maxpool layer with a Softmax function are followed to predict attentive weight for each neighbor point. The motion feature $\tilde{m}_{t}^l$ for a point in frame $t$ can be represented as the weighted-sum of the motion features of the neighbor points in frame $t-1$. This operation will be applied on all points in $X_{t}^l$ to generate motion features $\tilde{M}_{t}^l$. Intuitively, the attention mechanism can assign higher weights to the neighbor points that are more likely to be corresponding to the center point $x_{t}^l$. As a result, the proposed motion align module can reasonably estimate the motion features of frame $t$.

\subsubsection{Loss}
We adopt Chamfer Distance (CD) between the ground truth point cloud $P_t\in \mathbb{R}^{N\times 3}$ and the predicted point cloud $\hat{P}_t\in \mathbb{R}^{N\times 3}$ as the loss function. Chamfer Distance \cite{fan2017point} is a commonly used metric to measure the similarity between two point clouds, which can be formulated as:

\begin{equation}
\label{eq:cd}
    \mathcal{L}=\frac{1}{N}\sum_{\hat{x}^i\in\hat{P}_t}\min_{x^j\in P_t}\left\|\hat{x}^i-x^j\right\| + \frac{1}{N}\sum_{x^j\in P_t}\min_{\hat{x}^i\in \hat{P}_t}\left\|\hat{x}^i-x^j\right\|
\end{equation}

\section{Experiments}
  
\subsection{Datasets}
The proposed method is evaluated on two large scale outdoor LiDAR datasets, namely KITTI odometry dataset \cite{geiger2012we} and Argoverse dataset \cite{chang2019argoverse}. KITTI dataset contains 11 sequences (00-11) with ground truth, and we use sequence 00 to 05 to train the network, 06 to 07 to validate and 08 to 10 to test. We use the 3D tracking branch in Argoverse dataset and follow the default splits to train, validate and test the networks. For each dataset, we sample 10 consecutive point clouds as a sequence, where the first 5 frames are the input point clouds and the last 5 frames are used as the ground truth of predicted point clouds.

\subsection{Baseline methods}
We compare our method with several baseline methods to demonstrate the performance. (1) PointRNN (LSTM) and PointRNN (GRU): PointRNN is proposed in \cite{fan2019pointrnn} and we call the two variations using LSTM and GRU as PointRNN (LSTM) and PointRNN (GRU). The networks are re-trained on the two datasets due to the different preprocessing of point clouds in \cite{fan2019pointrnn}. (2) PointNet++ (LSTM) and PointNet++ (GRU): We define two additional baselines based on the encoder-decoder architecture of PointNet++ \cite{qi2017pointnet++}. Firstly the input point clouds are encoded into one-dimensional global features. After that, LSTM or GRU is utilized to model the temporal information and the generated global feature is decoded into the predicted per-point motions to generate future point clouds. (3) Scene flow: We utilize the 3D scene flow estimation network FlowNet3D \cite{liu2019flownet3d} to estimate the 3D scene flow between the last two input point clouds and use the estimated scene flow to predict the future point clouds. FlowNet3D is finetuned on the two datasets to achieve better performance.

\subsection{Implementation details}
Noting that we provide two versions of MotionRNN named MotionLSTM and MotionGRU, and MoNet using the two versions are denoted as MoNet (LSTM) and MoNet (GRU), respectively. We stack multi-layer MECell and MICell to construct a hierarchical structure and we use a 3-layer architecture in our implementation as shown in Fig.~\ref{fig:network}. The points and states of last frame for the first MECell are initialized to zero and the input content features of the first layer (\emph{i.e.}, the input point clouds) are also initialized to zero. The network is implemented using PyTorch \cite{paszke2019pytorch} and Adam \cite{kingma2014adam} is used as the optimizer. The point clouds are downsampled to 16384 points using random sampling during training. The network is trained and evaluated on NVIDIA GeForce RTX 2080Ti. More details about the network structure are provided in the supplementary materials.

\begin{figure*}[htbp]
    \centering
    \includegraphics[width=0.99\textwidth]{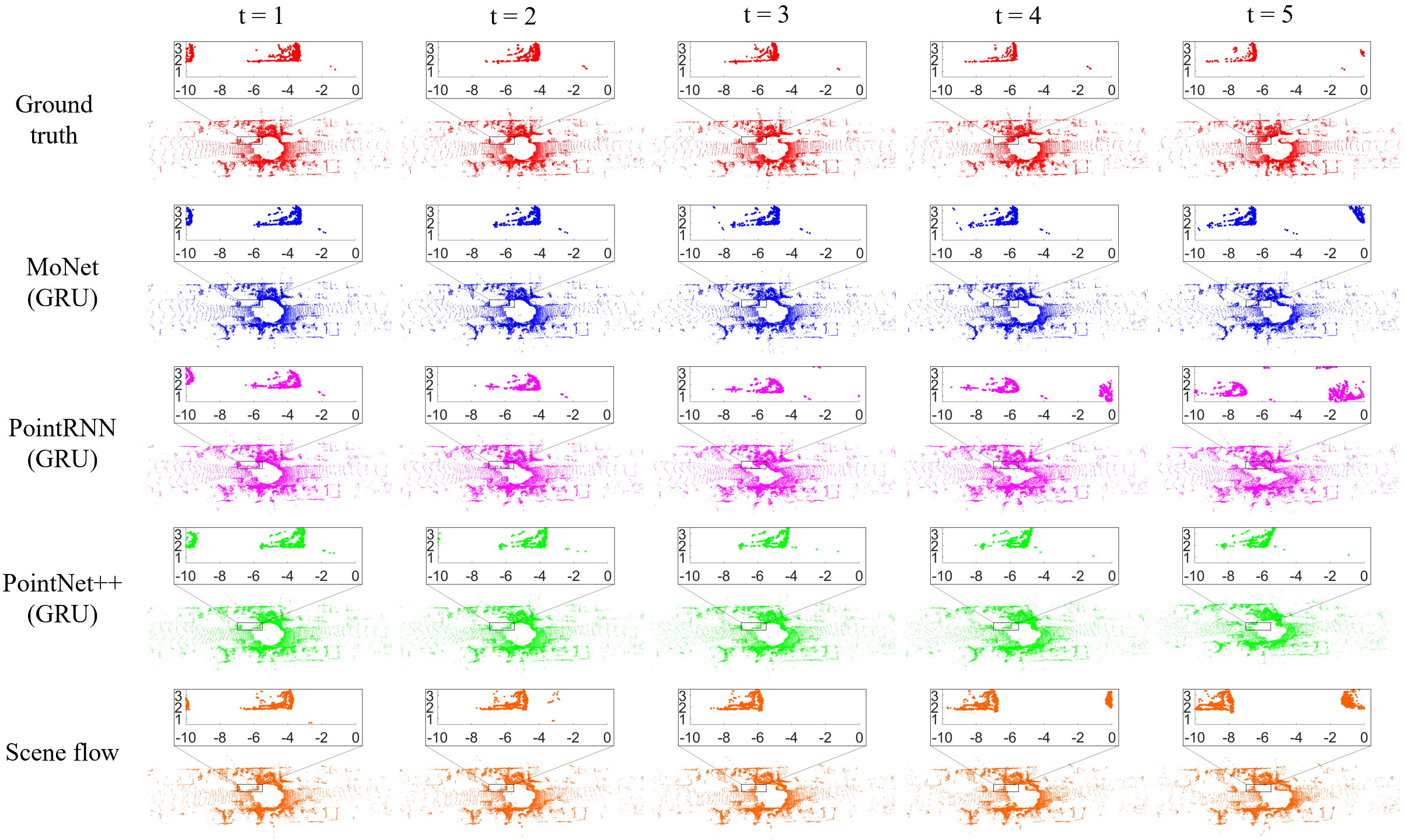}
    \caption{Qualitative visualization of predicted point clouds of different methods on KITTI odometry dataset. From left to right are the predicted future point clouds from $t=1$ to $t=5$. Rows from top to bottom display the ground truth point clouds and the predicted results of MoNet (GRU), PointRNN (GRU), PointNet++ (GRU) and Scene flow.}
    \label{fig:qualitative}
    \vspace{-3mm}
\end{figure*}

\subsection{Qualitative evaluation}
We provide several qualitative evaluation results on KITTI odometry dataset in Fig.~\ref{fig:qualitative} and the point clouds are downsampled to 32768 points during pre-processing. The images from left to right display the future point clouds from frame $t=1$ to $t=5$ and different rows represent the results of different methods. For a fair comparison, all of the methods using recurrent neural network are based on GRU. Besides, we zoom in an area containing the point cloud of a vehicle for better visualization. According to Fig.~\ref{fig:qualitative}, the point clouds predicted by our MoNet (GRU) are more consistent with the ground truth point clouds than other baseline methods. From the zoomed-in point clouds, the proposed method precisely predicts the position of the vehicle and the geometry of the point cloud is well preserved. Compared with our method, the point clouds of the vehicle from PointNet++ (GRU) are deformed, which is due to the poor representative ability of one-dimensional global features. PointRNN (GRU) can not correctly predict the position of the vehicle due to the lack of explicit modeling of motion information. Based on the qualitative results, the proposed method can precisely predict the motion of the point clouds and well preserve the details of point clouds.

\subsection{Quantitative evaluation}
\subsubsection{Evaluation metrics}
We adopt two metrics to evaluate the performance, namely Chamfer Distance (CD)\cite{fan2017point} and Earth Mover's Distance (EMD) \cite{rubner2000earth}. CD is described previously in Eq~\ref{eq:cd}. EMD is also a commonly used metric to compare two point clouds, which is implemented by solving a linear assignment problem between two point clouds. Given two point clouds $P_t\in\mathbb{R}^{N\times3}$ and $\hat{P}_t\in\mathbb{R}^{N\times3}$, EMD can be calculated as:
\begin{equation}
    EMD=\min_{\phi:\hat{P_t}\rightarrow P_t}\frac{1}{N}\sum_{\hat{x}^j\in \hat{P}_j}\left\|\hat{x}-\phi(\hat{x})\right\|
\end{equation}
where $\phi:\hat{P}_t\rightarrow P_t$ is a bijection from $\hat{P}_t$ to $P_t$. Due to the high computational complexity of EMD, the point clouds are downsampled to 16384 using random sampling during pre-processing in quantitative evaluation. 

\begin{figure*}[!t]
	\centering
	\subfigure[KITTI odometry dataset (LSTM)]{
		\includegraphics[width=0.235\textwidth]{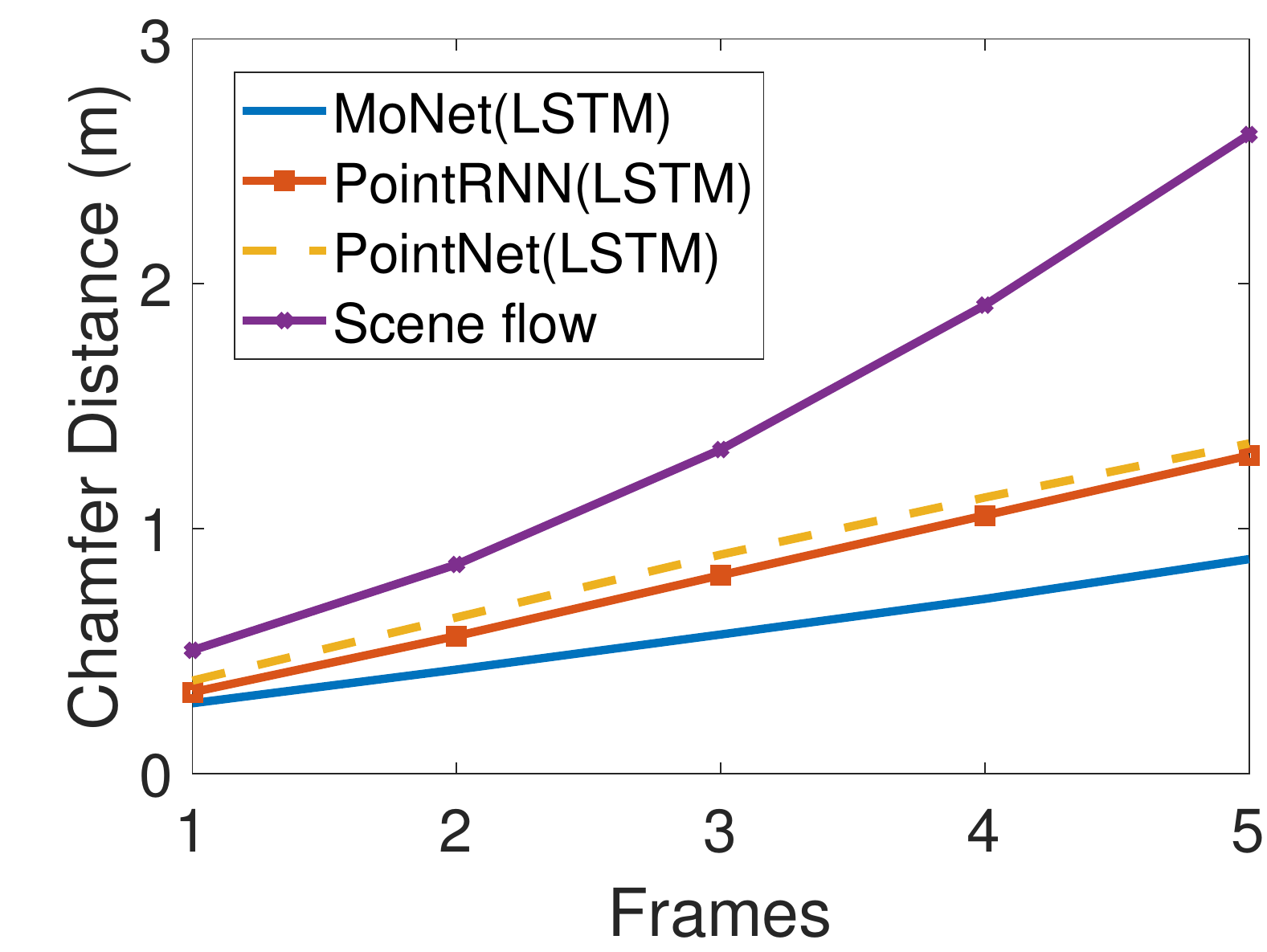}
	}
	\subfigure[Argoverse dataset (LSTM)]{
	\includegraphics[width=0.235\textwidth]{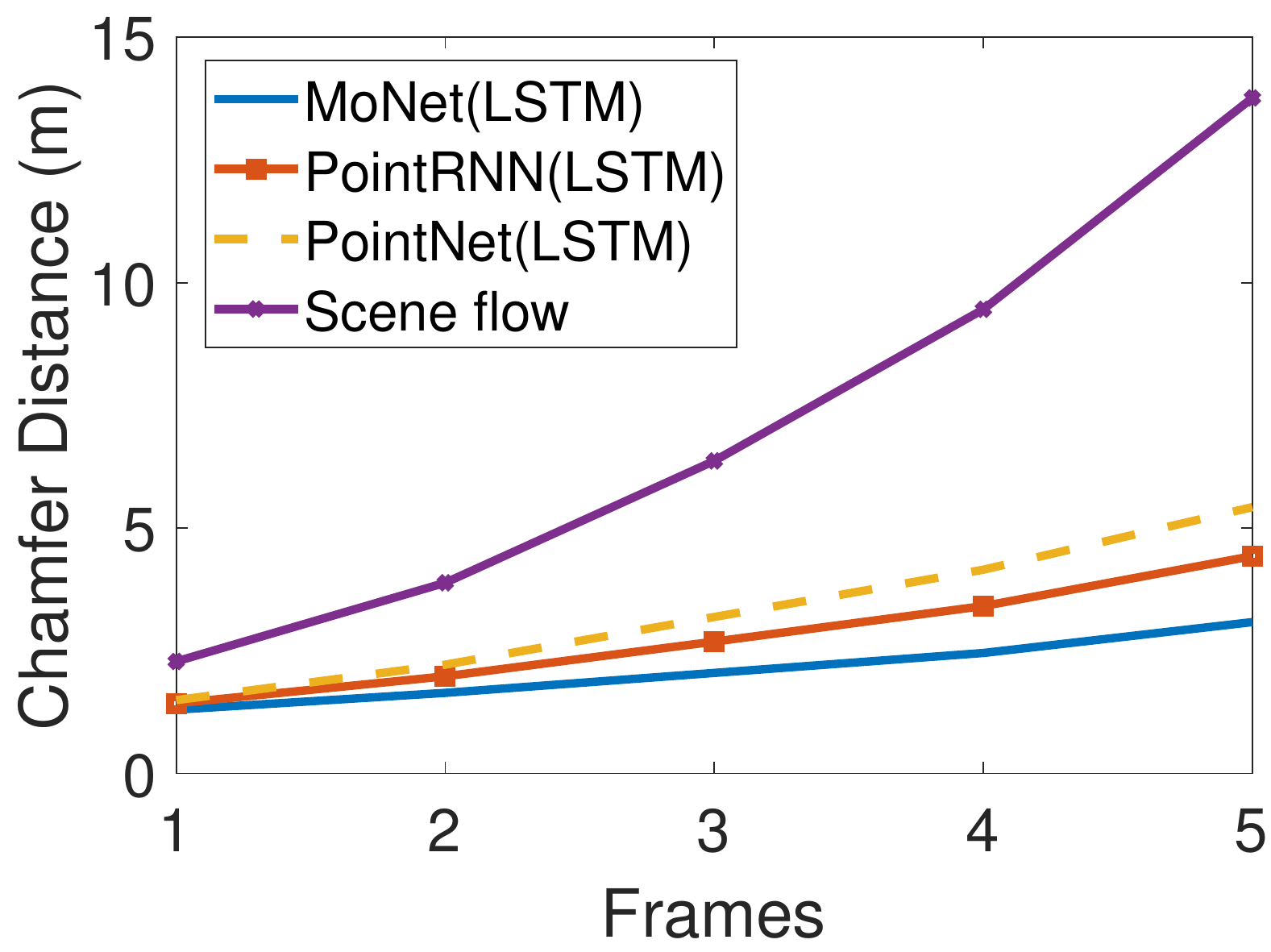}
	}
	\subfigure[KITTI odometry dataset (GRU)]{
		\includegraphics[width=0.235\textwidth]{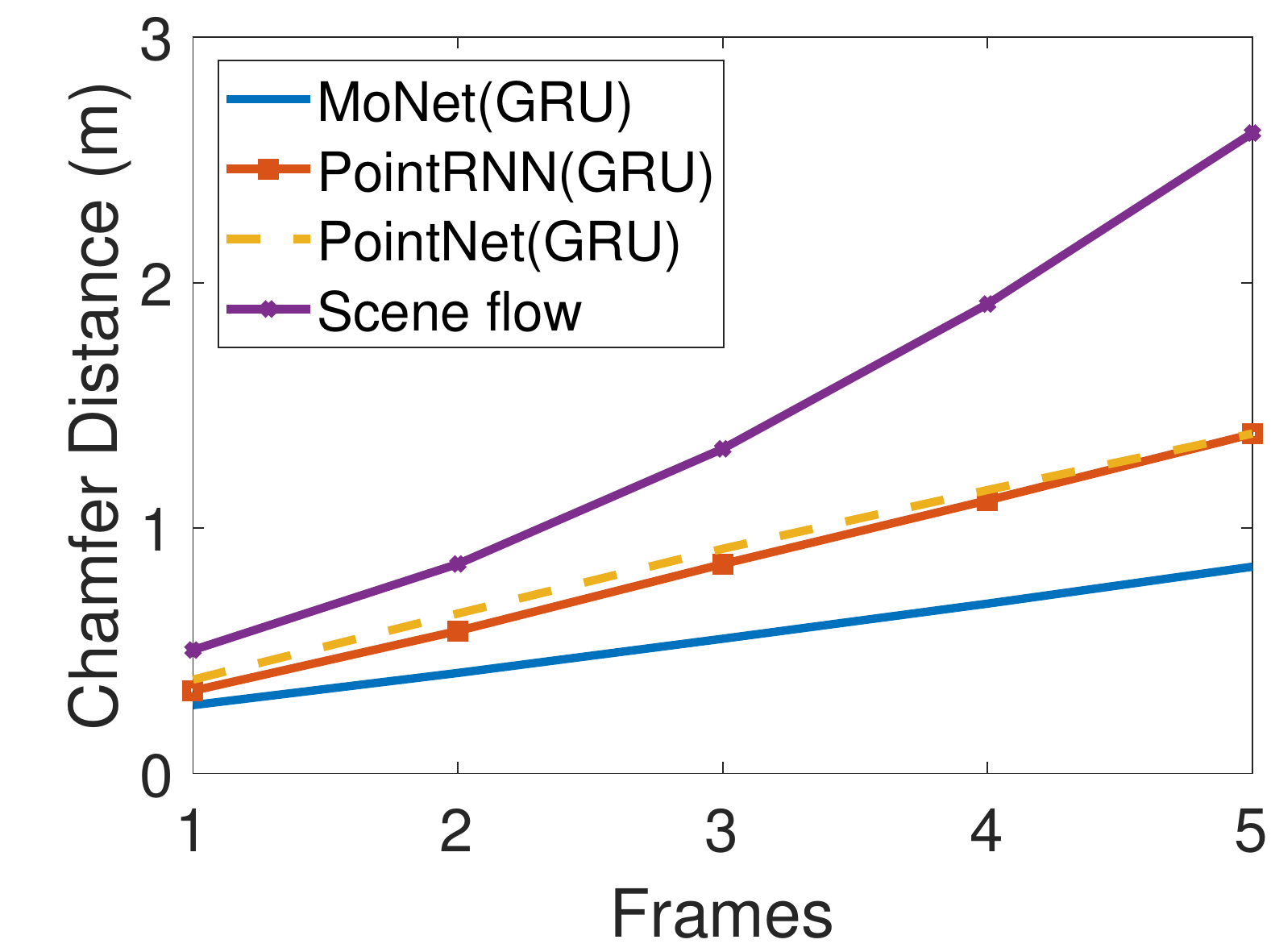}
	}
	\subfigure[Argoverse dataset (GRU)]{
		\includegraphics[width=0.235\textwidth]{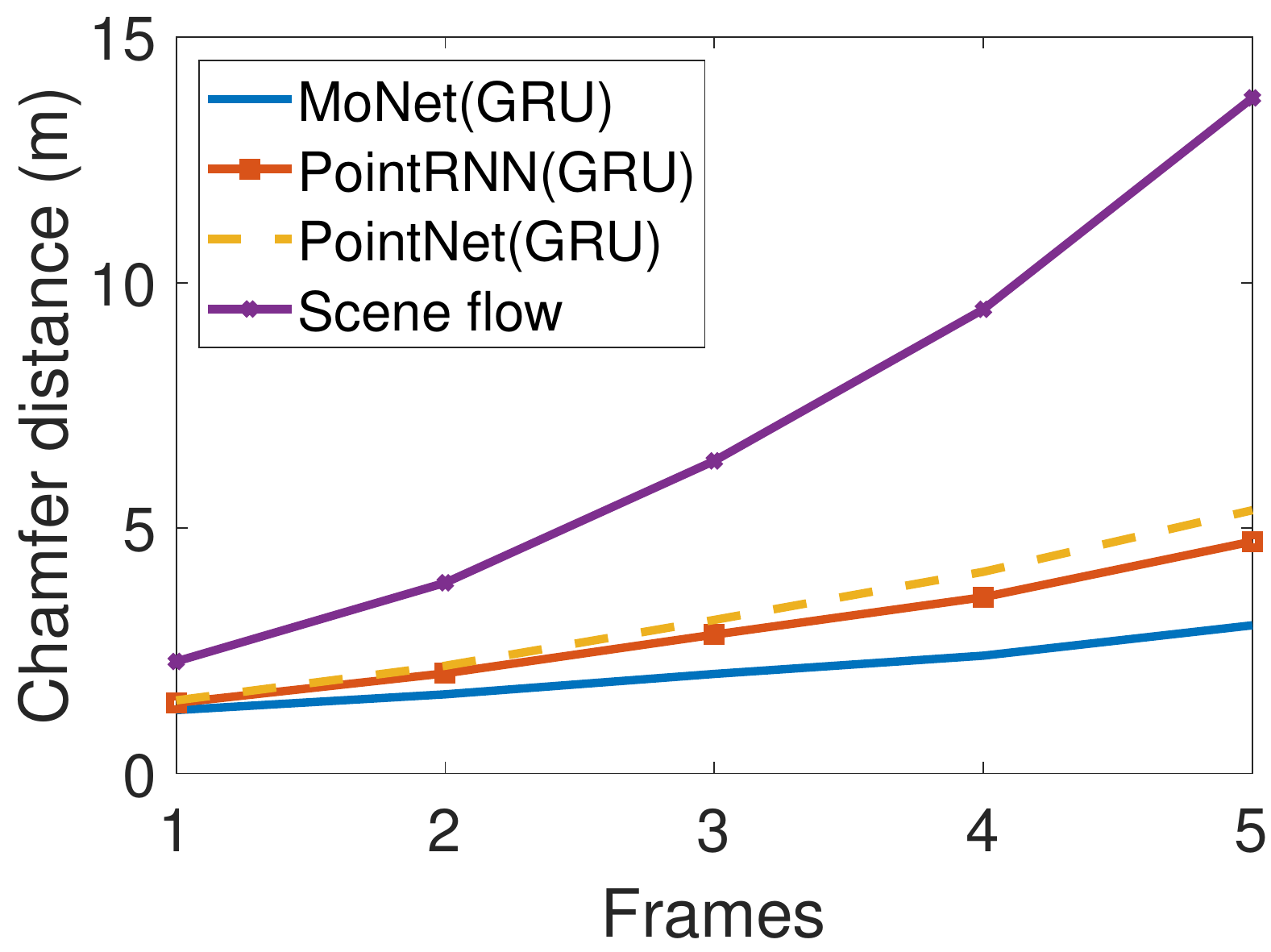}
	}
    \caption{Chamfer Distance of the proposed methods and baseline methods on KITTI odometry dataset and Argoverse dataset. The left two images display the results using LSTM and the right two images show the results using GRU.}
	\label{fig:chamfer}
	\vspace{-3mm}
\end{figure*}

\begin{figure*}[!t]
	\centering
	\subfigure[KITTI odometry dataset (LSTM)]{
		\includegraphics[width=0.235\textwidth]{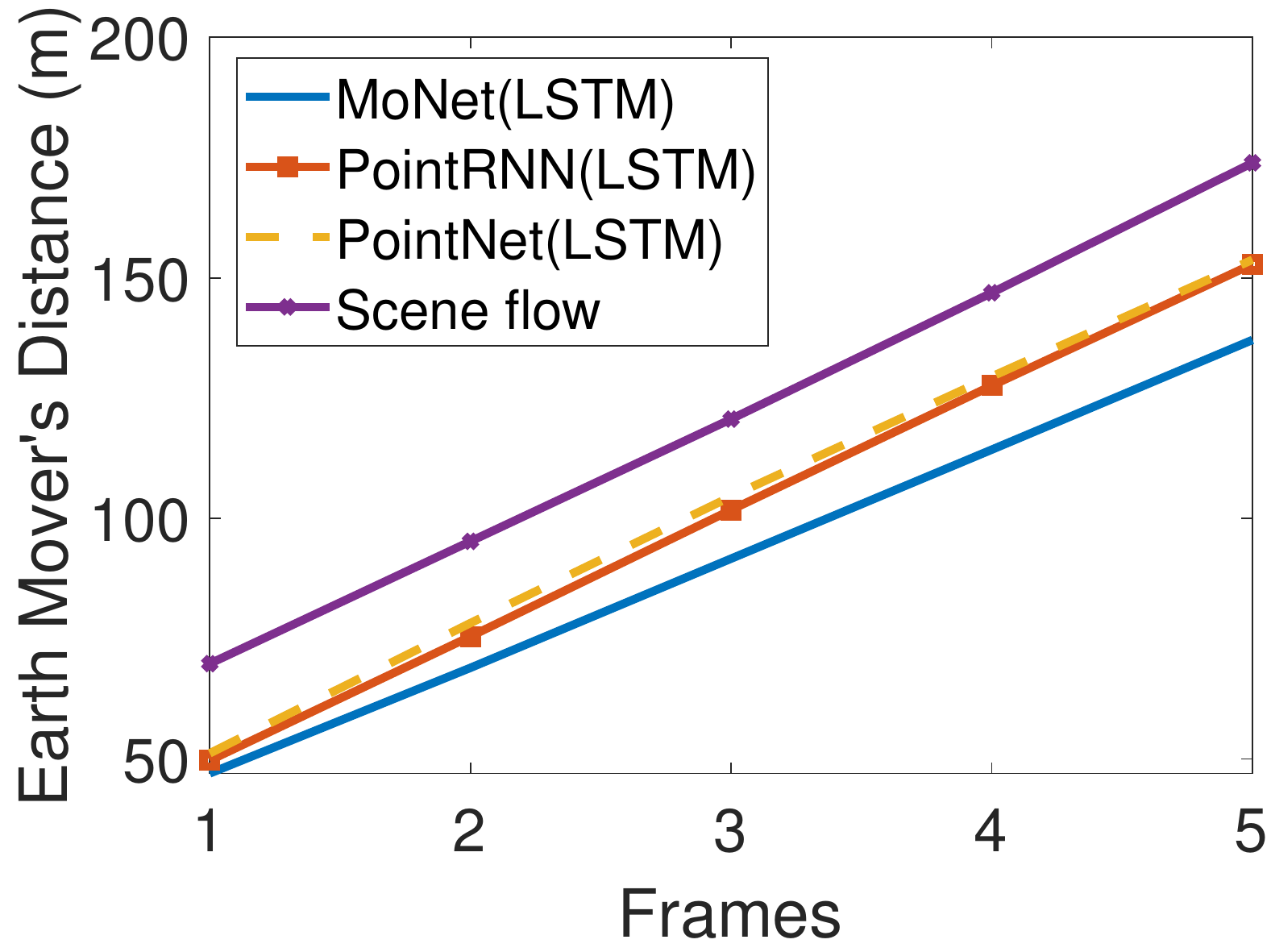}
	}
	\subfigure[Argoverse dataset (LSTM)]{
	\includegraphics[width=0.235\textwidth]{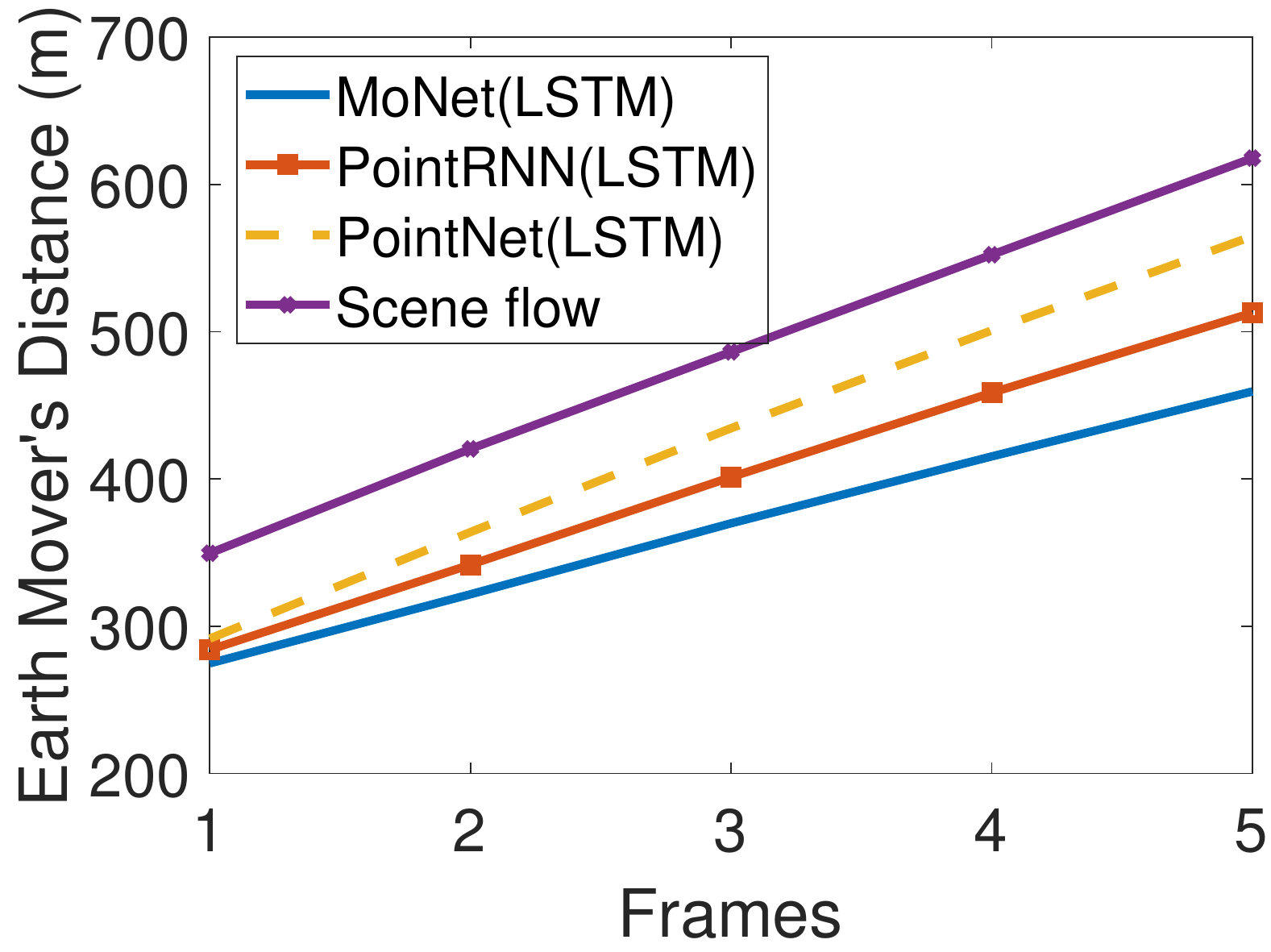}
	}
	\subfigure[KITTI odometry dataset (GRU)]{
		\includegraphics[width=0.235\textwidth]{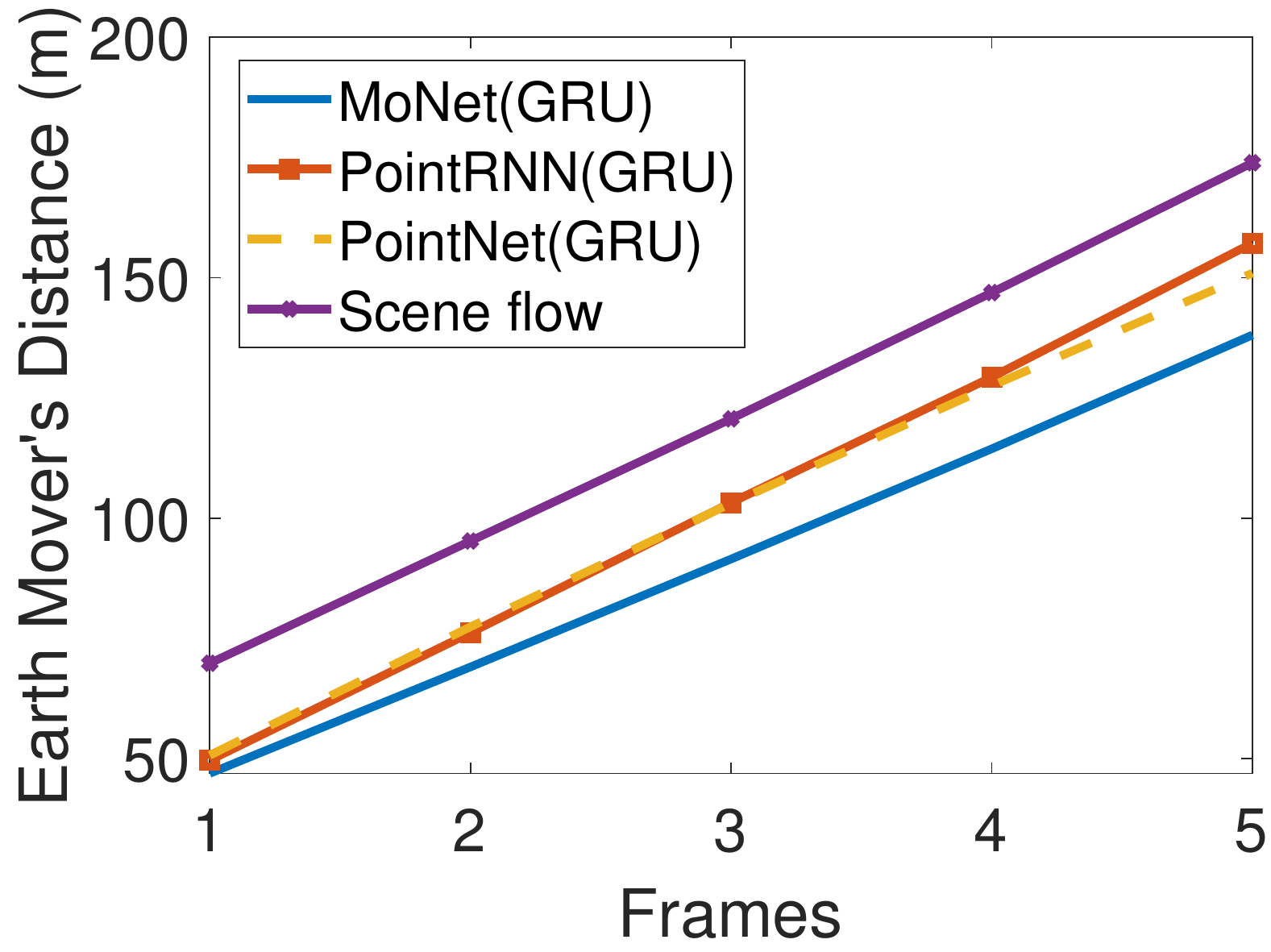}
	}
	\subfigure[Argoverse dataset (GRU)]{
		\includegraphics[width=0.235\textwidth]{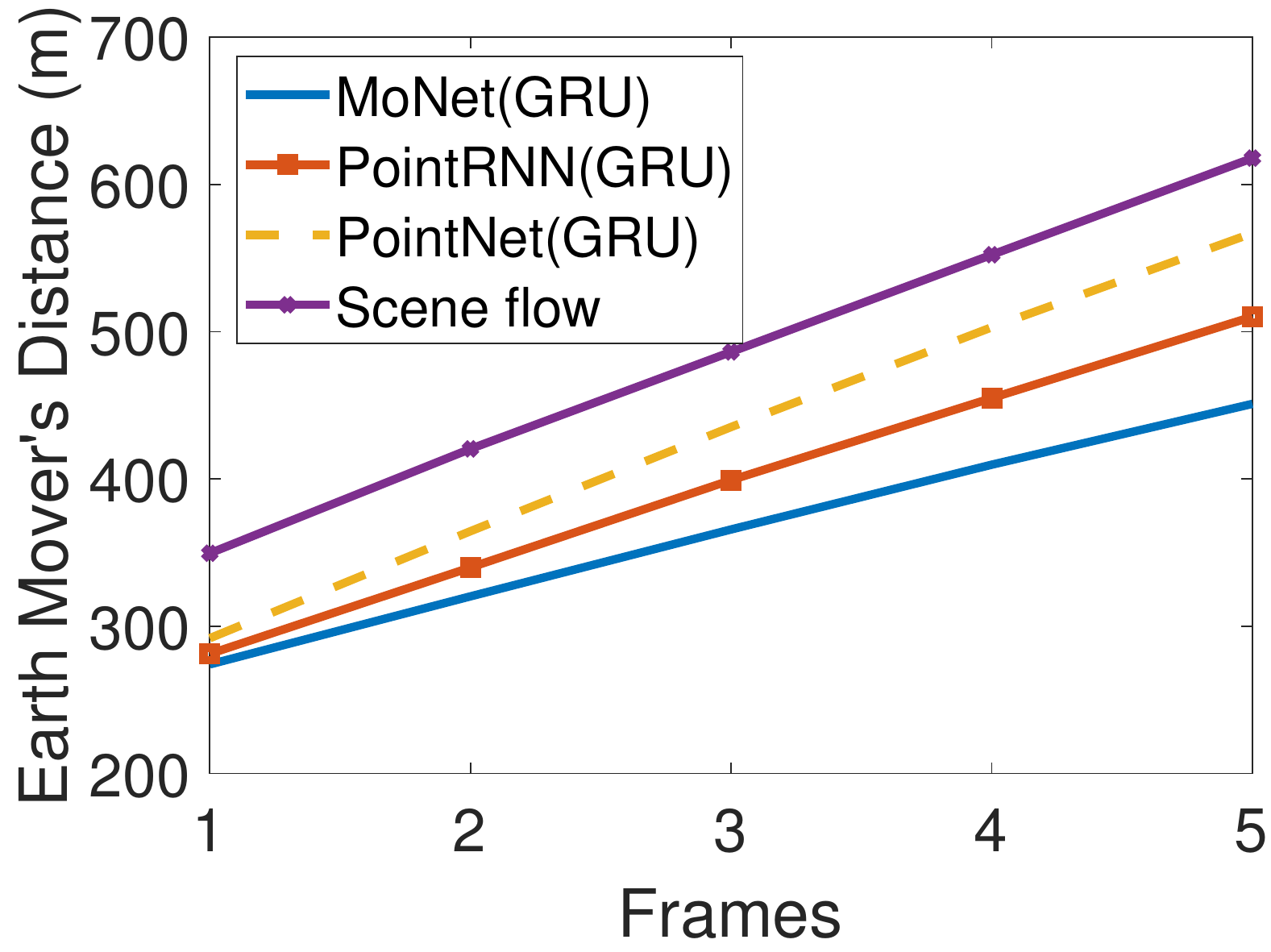}
	}
    \caption{Earth Mover's Distance (EMD) of the proposed methods and baseline methods on KITTI odometry dataset and Argoverse dataset. The left two images display the EMD using LSTM and the right two images show the results using GRU.}
	\label{fig:emd}
	\vspace{-3mm}
\end{figure*}

\subsubsection{Chamfer Distance}
We calculate the Chamfer Distance (CD) between predicted point clouds and ground truth ones on 5 future frames on KITTI and Argoverse dataset and display the results in Fig.~\ref{fig:chamfer}. For better comparison, we display the results using LSTM and GRU on the left two images and right two images in Fig.~\ref{fig:chamfer}, respectively. The CD of our methods is lower than other baseline methods by an obvious margin on two datasets. For example, the average CD on KITTI dataset of MoNet (LSTM) is about 30\% smaller than the best baseline PointRNN (LSTM). The Chamfer Distance increases as the number of predicted frames increases, which is due to the increase in uncertainty. However, the growth of CD of our method is much slower than other methods, which demonstrates the robustness of the proposed method.

\subsubsection{Earth Mover's Distance}
Similar to Chamfer Distance, Earth Mover's Distance (EMD) on two datasets are shown in Fig.~\ref{fig:emd}. Compared with CD, EMD is more sensitive to the local details and the density distribution of the point clouds \cite{liu2020morphing}. According to Fig.~\ref{fig:emd}, the proposed MoNet significantly outperforms other baseline methods on this metric. For example, the EMD of the 5th frame on Argoverse dataset of our MoNet (GRU) is about 8\% smaller than PointRNN (GRU) and 15\% smaller than PointNet++ (GRU). The lower EMD indicates that our methods preserve the local details of point clouds better, which is also verified by the qualitative experiments results.

\begin{table}
    \centering
    \caption{Computation time (ms) for predicting 5 frames with different number of points.}
    \begin{tabular}{ccccccccc}
    \toprule
    Number of points & 16384 & 32768 & 65536 \\
    \midrule
    MoNet (GRU) & 91 & 129 & 280 \\
    PointRNN (GRU) & 89 & 125.5 & 268.7 \\
    PointNet++ (GRU) & 48 & 86 & 233 \\
    Scene flow & 64 & 115 & 216 \\
    \bottomrule
    \end{tabular}
    \label{tab:runtime}
    \vspace{-4mm}
\end{table}

\begin{table}
    \centering
    \caption{Precision, average Euclidean distance (AED) and average heading error (AHE) of 3D object detection results from predicted point clouds of different methods.}
    \begin{tabular}{cccc}
    \toprule
    Method & Precision & AED (m) & AHE ($^\circ$) \\
    \midrule
    MoNet (GRU) & \textbf{0.763} & \textbf{0.173} & \textbf{2.267} \\
    PointRNN (GRU) & 0.498 & 0.232 & 2.862 \\
    PointNet++ (GRU) & 0.458 & 0.237 & 3.039 \\
    Scene flow & 0.444 & 0.231 & 2.887 \\
    \bottomrule
    \end{tabular}
    \label{tab:3dDet}
    \vspace{-4mm}
\end{table}

\subsection{Efficiency}
Noting that the content features and motion features of previous frames do not need to be recalculated when a new point cloud is generated by the LiDAR. The computation time for predicting 5 future frames with 16384, 32768 and 65536 points is shown in Table~\ref{tab:runtime}. According to the results, the proposed method has a similar computation time with PointRNN, however, achieves much better performance. Specifically, the computation time of our method is 280 ms even with 65536 points, which is lower than the time required for typically LiDAR (10 Hz) to generate 5 frames.

\subsection{Applications}
The predicted point clouds can be applied to many applications like 3D object detection and semantic segmentation. Taken 3D object detection as an example, we adopt PV-RCNN \cite{shi2020pv} to detect cars from the original point clouds and the predicted ones. We use a metric named precision to evaluate the quality of the predicted point clouds. A car is detected if the predicted score is larger than a threshold $\tau_s=0.9$. For each detected car in the original point cloud, we search for the nearest detection in the predicted point clouds and if the 2D Euclidean distance is within a threshold $\tau_d=0.5 m$ and the heading error is within $\tau_{\theta} = 10^\circ$, the detection in the predicted point clouds is considered as valid and the precision is defined as the ratio of the valid detections. Besides, we also calculate the average Euclidean distance (AED) and average heading error (AHE) of valid detections to evaluate the accuracy of the predicted point clouds. The experiment is performed on Sequence 08 of KITTI dataset and the number of point clouds is set to 65536. The results are shown in Table~\ref{tab:3dDet}. According to Table~\ref{tab:3dDet}, the average precision across 5 frames of the proposed method is 0.763, which outperforms other baseline methods by a significant margin and the results also demonstrate the high consistency between the predicted point clouds and the original ones. Besides, the low AED and AHE show that the valid detections in the predicted point clouds are close to the original ones. Visualizations of 3D object detection are displayed in the supplementary material. The experiments on 3D object detection show that the predicted point clouds can be used for further perception and indicate the great application potential of the proposed method.

\subsection{Ablation study}
\subsubsection{MotionLSTM or MotionGRU?}
Noting that we provide two versions of MotionRNN (\emph{i.e.}, MotionLSTM and MotionGRU) to model the temporal correlations. To compare the performance of the two versions, we calculate the average Chamfer Distance (CD) and Earth Mover's Distance (EMD) of 5 future point clouds of MoNet (LSTM) and MoNet (GRU) and display the results in Table~\ref{tab:lstm_gru}. According to the results, the performance of MoNet (LSTM) and MoNet (GRU) is highly similar, which indicates that the proposed method is not sensitive to different recurrent neural network architectures. The performance of MoNet (GRU) is slightly better than MoNet (LSTM). Besides, the network parameters of MotionGRU is less than that of MotionLSTM, which also results in a faster inference speed and lower memory usage. Overall, MoNet (GRU) can be a better choice for point cloud prediction.

\begin{table}
    \centering
    \caption{Average CD (m) and EMD (m) of MoNet (LSTM) and MoNet (GRU) on two datasets.}
    \begin{tabular}{cccc}
    \toprule
        Dataset & Model & CD & EMD \\
    \midrule
        \multirow{2}{*}{KITTI}  & MoNet (LSTM) & 0.573 & \textbf{91.79} \\
        ~ & MoNet (GRU) & \textbf{0.554} & 91.97 \\
    \midrule
        \multirow{2}{*}{Argoverse} & MoNet (LSTM) & 2.105 & 368.21 \\
        ~ & MoNet (GRU) & \textbf{2.069} & \textbf{364.14} \\
    \bottomrule
    \end{tabular}
    \label{tab:lstm_gru}
\end{table}

\subsubsection{Motion features and content features}

\begin{figure}
    \centering
    \includegraphics[width=0.46\textwidth]{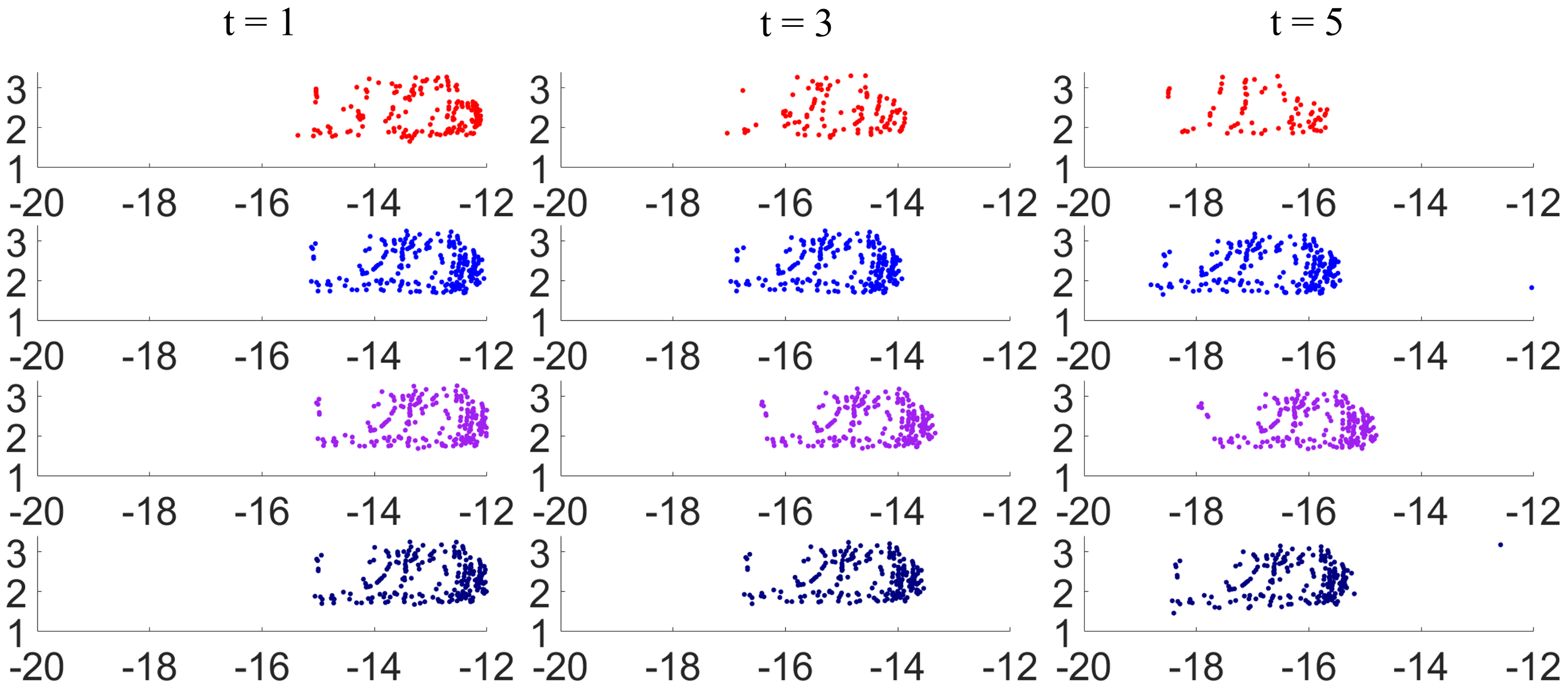}
    \caption{From left to right are future frames $t=1,3,5$. From top to bottom are results of ground truth, full MoNet (GRU), model w/o motion features and w/o content features.}
    \label{fig:moco}
\end{figure}

As we claimed before, motion features and content features can both contribute to the prediction of future point clouds. To demonstrate the significance of the combination of motion and content features, we remove the motion features and content features separately to compare the performance. The rows from top to bottom of Fig.~\ref{fig:moco} display the point clouds of a vehicle from ground truth point cloud, the results of full MoNet (GRU) and the results of the model without motion and content features. The model without motion features results in biased motion estimation. The model without content features can correctly predict the motion, however, the point clouds are slightly deformed. The average CD and EMD of MoNet (GRU) with and without motion features and content features are shown in Table~\ref{tab:motion_content}. According to the results, the combination of content and motion features significantly improves the performance of the proposed MoNet. For example, the CD without motion features and content features are 0.092 m and 0.083 m higher than full MoNet (GRU) on KITTI dataset, respectively. Besides, the EMD of the full model is also lower than the model without motion or content features. Overall, the combination of content and motion features leverages the strength of both features to obtain precise motion estimation and better preservation of local details.

\begin{table}
    \centering
    \caption{Average CD (m) and EMD (m) of MoNet (GRU), without (w/o) motion features and w/o content features on two datasets.}
    \begin{tabular}{cccc}
    \toprule
    Dataset & Options & CD & EMD \\
    \midrule
    \multirow{3}{*}{KITTI}  & full model & \textbf{0.554} & \textbf{91.97} \\
    ~ & w/o motion features & 0.646 & 92.98 \\
    ~ & w/o content features & 0.637 & 93.83 \\
    \midrule
    \multirow{3}{*}{Argoverse} & full model & \textbf{2.069} & \textbf{364.14} \\
    ~ & w/o motion features & 2.200 & 372.48 \\
    ~ & w/o content features & 2.158 & 366.40 \\
    \bottomrule
    \end{tabular}
    \label{tab:motion_content}
\end{table}

\section{Conclusion}
In this paper, we explore a problem named Point Cloud Prediction, which aims to predict future frames given past point cloud sequence. To achieve that, we propose a novel motion-based neural network named MoNet. Specifically, the proposed MoNet integrates motion features into the prediction pipeline and combines that with content features. Besides, a recurrent neural network named MotionRNN is proposed to capture the temporal correlations of both features across point cloud sequence and a novel motion align module is adopted to estimate motion features without future point cloud frames. Both qualitative and quantitative experiments are performed on KITTI odometry dataset and Argoverse dataset to demonstrate the performance of the proposed method. Abundant ablation studies show that the combination of motion and content features enables the model to precisely predict the motions and also well preserve the structures. Moreover, experiments on applications reveal the practical potential of the proposed method.

{\small
\bibliographystyle{ieee_fullname}
\bibliography{egbib}
}

\end{document}